\documentclass[conference]{IEEEtran}

\makeatother
\usepackage{amsmath,amssymb,amsfonts}
\usepackage{mathtools}
\usepackage{graphicx}
\usepackage{subcaption} 
\usepackage{booktabs}
\usepackage{multirow}
\usepackage{hyperref}
\usepackage{xcolor}
\usepackage{url}
\usepackage{siunitx}
\usepackage{microtype}
\usepackage{enumitem} 
\usepackage{bm} 
\usepackage{fontspec} 
\newfontface\banglafont[Script=Bengali]{Kalpurush.ttf}
\newcommand{\bn}[1]{{\banglafont #1}}

\begin {document}

\title{
\begin{flushleft}
\fontsize{9}{11}\selectfont
2025 28th International Conference on Computer and Information Technology (ICCIT)\\
19--21 December 2025, Cox's Bazar, Bangladesh
\end{flushleft}

\vspace{0.15cm}

\begin{center}
\Large
When a Nation Speaks: Machine Learning and NLP in \\
People’s Sentiment Analysis During Bangladesh’s 2024 Mass Uprising
\end{center}
}
\vspace{-0.6cm}

\author{
\IEEEauthorblockN{
Md. Samiul Alim\textsuperscript{1},
Mahir Shahriar Tamim\textsuperscript{1},
Maisha Rahman\textsuperscript{1},
Tanvir Ahmed Khan\textsuperscript{1},
Md Mushfique Anwar\textsuperscript{2}
}

\IEEEauthorblockA{
\textsuperscript{1}North South University, Dhaka 1229, Bangladesh \\
\{samiul.alim01, mahir.shahriar, maisha.rahman12, khan.tanvir01\}@northsouth.edu
}

\IEEEauthorblockA{
\textsuperscript{2}Jahangirnagar University, Dhaka, Bangladesh \\
manwar@juniv.edu
}
}

\maketitle


\begin{abstract}
Sentiment analysis, an emerging research area within natural language processing (NLP), has primarily been explored in contexts like elections and social media trends, but there remains a significant gap in understanding emotional dynamics during civil unrest, particularly in the Bangla language. Our study pioneers sentiment analysis in Bangla during a national crisis by examining public emotions amid Bangladesh's 2024 mass uprising. We curated a unique dataset of 2,028 annotated news headlines from major Facebook news portals, classifying them into \textit{Outrage}, \textit{Hope}, and \textit{Despair}. Through Latent Dirichlet Allocation (LDA), we identified prevalent themes like political corruption and public protests, and analyzed how events such as internet blackouts shaped sentiment patterns. It outperformed multilingual transformers (mBERT: 67\%, XLM-\hspace{0pt}RoBERTa: 71\%) and traditional machine learning methods (SVM and Logistic Regression: both 70\%). These results highlight the effectiveness of language-specific models and offer valuable insights into public sentiment during political turmoil.
\end{abstract}

\begin{IEEEkeywords}
Sentiment Analysis,
Natural Language Processing\hspace{0pt} (NLP),
Bangla language,
Mass Uprising,
Latent Dirichlet Allocation\hspace{0pt} (LDA),
Machine Learning\hspace{0pt} (ML).
\end{IEEEkeywords}

\section{Introduction}
In the digital era, social media has become a central space for public expression—especially during political unrest. Platforms like Facebook amplify citizens’ voices and serve as informal channels for news, emotional release, and collective mobilization. During national crises, online discourse is flooded with emotions such as outrage, hope, and despair. Analyzing these dynamics in real time offers valuable insight into public sentiment, media influence, and the trajectory of social movements.

Sentiment analysis has grown into a core application of Natural Language Processing (NLP), particularly in electoral forecasting \cite{r8}, brand monitoring \cite{r2}, and event analysis. Yet most existing work centers on binary sentiment or broad opinion trends, often in English and on structured platforms like Twitter. A gap remains in examining fine-grained emotional signals in low-resource languages during high-stakes political events.

In the Bangla NLP domain, datasets such as Motamot \cite{r1} and recent protest-news collections \cite{mosleh2023bangla} provide useful foundations. However, they are limited by: (1) their focus on routine political events rather than civil unrest; (2) reliance on binary or coarse sentiment labels, overlooking emotions like despair or hope; and (3) a lack of alignment with real-time events unfolding under censorship and political tension.

To address these gaps, we analyze the “July–August Mass Uprising of 2024” in Bangladesh\footnote{\url{https://en.wikipedia.org/wiki/July_Revolution_(Bangladesh)}}
, a period of intense upheaval, mass protests, and the eventual collapse of an authoritarian regime. With traditional media constrained by censorship, Facebook emerged as a key source of real-time updates and public discourse, especially among youth. This context provides a rare opportunity to study sentiment amid rapid emotional shifts and societal transformation.

We introduce a new Bangla dataset of Facebook-shared news headlines from the 2024 uprising. Each headline is manually annotated with a three-class emotion taxonomy—\textit{Outrage}, \textit{Hope}, and \textit{Despair}—informed by sociological theories of protest psychology \cite{jasper1998emotions}. A pilot study produced high inter-annotator agreement ($\kappa=0.78$), confirming the taxonomy’s reliability for capturing emotions in crisis reporting.

Our contributions are:  
\begin{itemize}[leftmargin=*,nosep]
    \item \textbf{Novel Dataset:} Collected and annotated 2,028 news headlines from prominent Bangladeshi Facebook news portals covering the July–August 2024 mass uprising—first dataset capturing crisis time Bangla sentiment.  
    \item \textbf{Granular Emotional Categories:} Introduced a three-class taxonomy \textit{Outrage}, \textit{Hope}, and \textit{Despair} enabling nuanced analysis beyond traditional binary sentiment labels.  
    \item \textbf{Insights into Crisis-Driven Sentiment:} Conducted temporal sentiment analysis revealing how events like internet blackouts, leadership changes, and natural disasters shaped public emotions during the uprising.  
\end{itemize}

\section{Related Work}
Prior work on political sentiment in Bangla and beyond spans dataset development, modeling, and event analysis. In Bangla, Motamot \cite{r1} provides 7,058 election-related instances using pre-trained and large language models, but relies mainly on binary labels and exhibits bias and variation issues. Protest-related news classification \cite{r3} (8,039 articles, eight classes) shows the strength of transformers, yet faces access restrictions, class imbalance, and oversampling concerns. Overall, existing Bangla resources rarely capture crisis-time discourse or fine-grained emotions such as despair and hope.

Broader social media research examines sentiment trends in elections and public opinion. Twitter-based favorability prediction \cite{r2} highlights challenges in preprocessing informal text and handling high-dimensional embeddings. Indian election sentiment using LSTM/SVM \cite{r7} suffers from small samples and imbalance, while Australian election sub-event detection \cite{unankard2014predicting} exposes demographic and geographic gaps. Aspect-based sentiment with RNN/LSTM \cite{r4} enhances granularity but increases computational cost and yields mixed accuracy, and dynamic sentiment work on U.S. political tweets \cite{r5} reports class imbalance and annotation bias.

Crisis-era sentiment analysis outside Bangla, such as Arabic tweets during the Tunisian uprising \cite{r00}, used SVM/Naïve Bayes but struggled with stopword limitations and weak generalization. Across these studies, common issues include coarse labels, dataset imbalance or access constraints, and domain shifts away from real crisis conditions. Our work addresses this gap through a Bangla crisis-time corpus and a three-emotion taxonomy—outrage, hope, and despair—tailored to protest contexts.
\section{Dataset and Methodology}

\begin{figure}[t]
    \centering
    \includegraphics[width=\columnwidth]{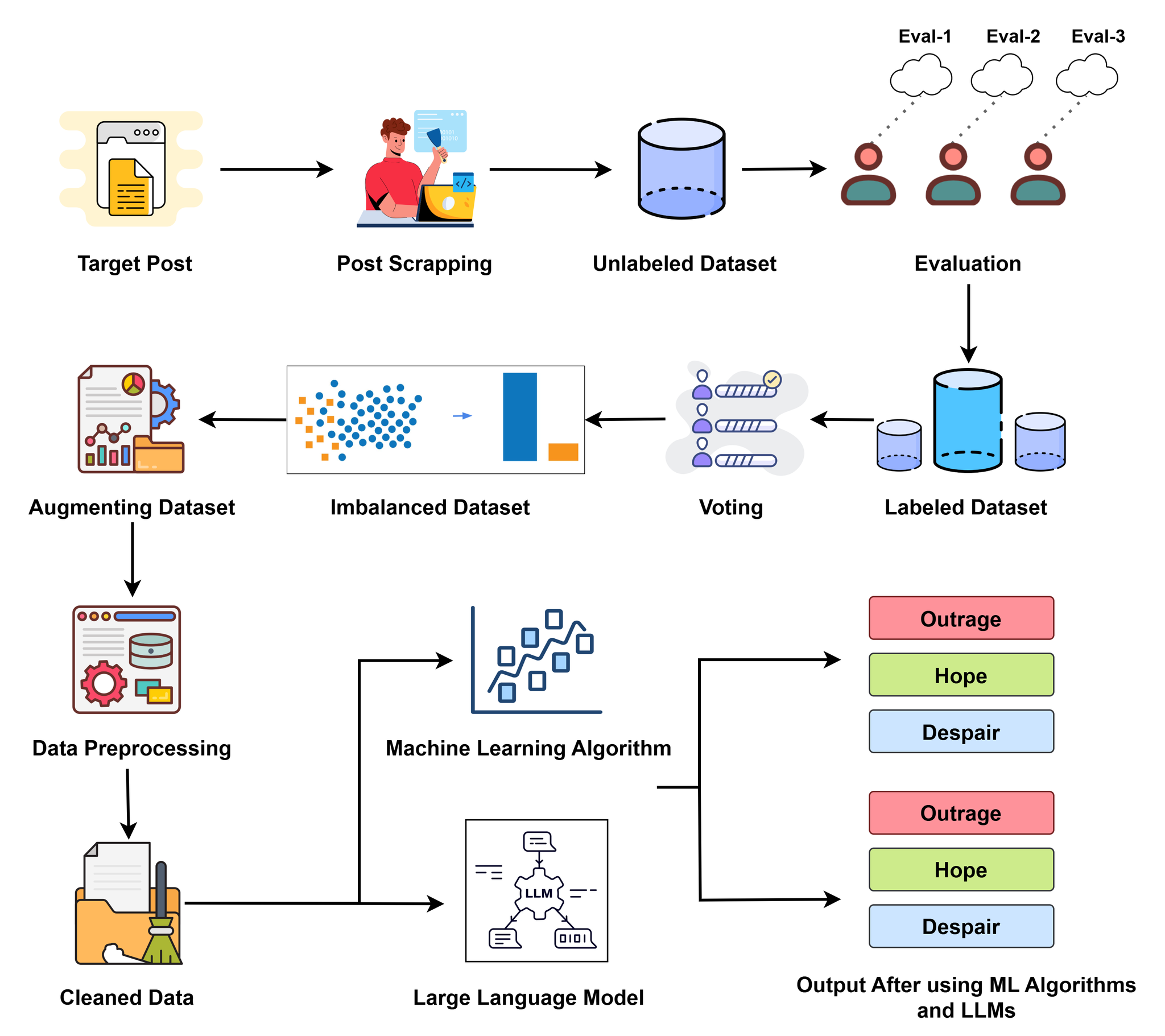} 
    \caption{Workflow for sentiment analysis, involving data collection, labeling, augmentation, preprocessing, and classification into Outrage, Hope, and Despair.}
    \label{fig:workflow}
\end{figure}

\subsection{Crisis-Centric Data Collection Framework}
\begin{figure}[t]
    \centering
    \includegraphics[width=\columnwidth]{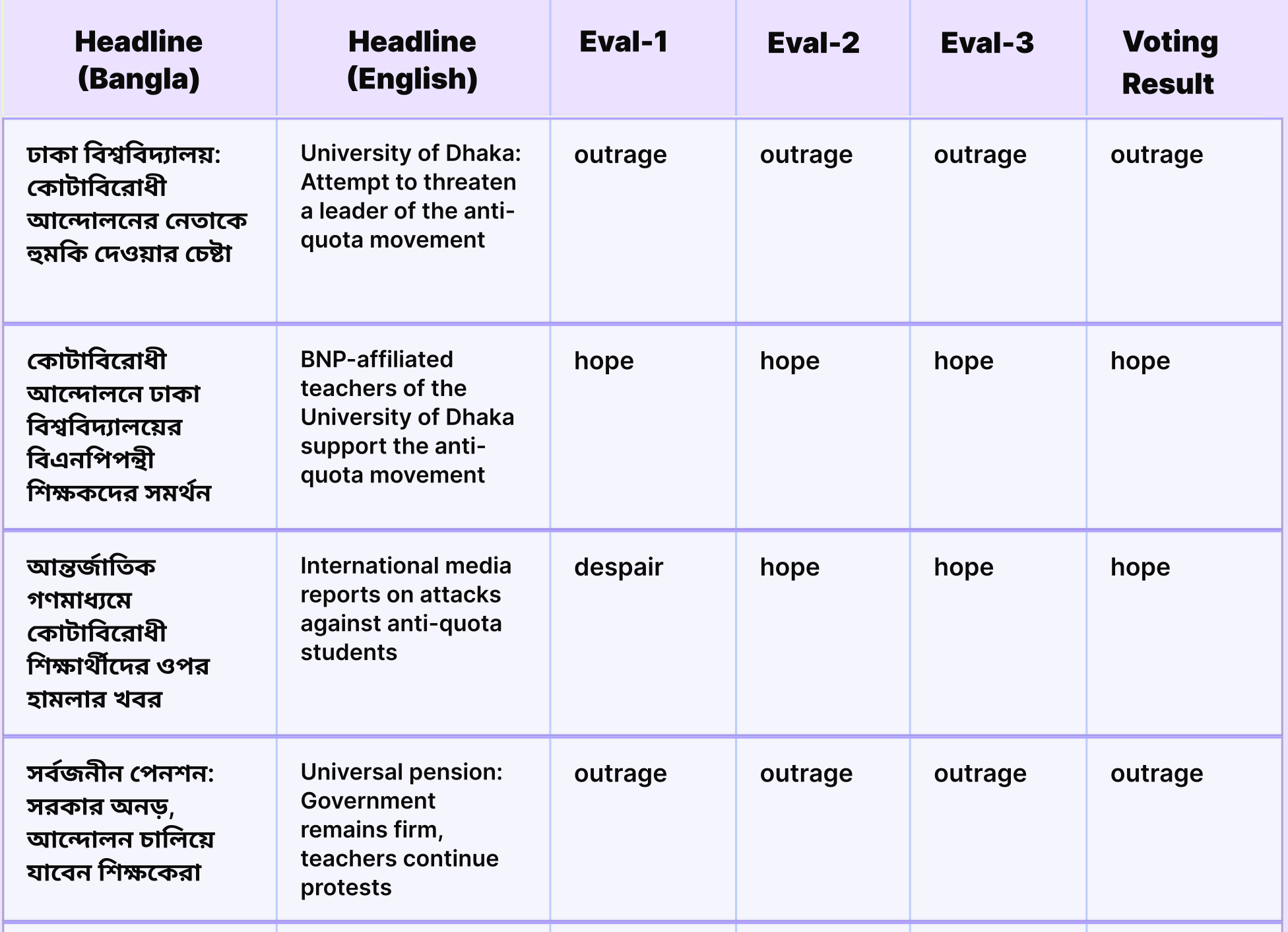} 
    \caption{Sample Comments with Sentiment Classification, Annotation, and Voting Outcome.}
    \label{fig:eval_table}
\end{figure}

We construct a temporally aligned corpus of crisis headlines from prominent Bangladeshi news outlets on Facebook, spanning July 5–August 30, 2024. Only uprising-related content is retained, with relevance scored by a function $\phi_{\text{rel}}(h)$ based on keywords and temporal alignment. The filtered dataset $\mathcal{D}_{\text{filtered}}$ is defined as:
\begin{equation}
    \mathcal{D}_{\text{filtered}} = \{\, h \in \mathcal{D}_{\text{raw}} : \phi_{\text{rel}}(h) > \tau \,\},
\end{equation}
where $\tau$ is the relevance threshold. Figure~\ref{fig:eval_table} shows sample comments and sentiment annotation outcomes. 

During the July 19–23 internet shutdown, we observe that outrage posts dropped significantly:
\begin{equation}
    P(\text{Outrage}\mid t \in \mathcal{T}_{\text{blackout}}) \ll P(\text{Outrage}\mid t \notin \mathcal{T}_{\text{blackout}}).
\end{equation}
This demonstrates the critical role of digital communication in shaping emotional responses during crises.

\subsection{Annotation Framework}

\begin{figure}[ht]
    \centering
    \begin{subfigure}{0.32\columnwidth}
        \centering
        \includegraphics[width=\linewidth]{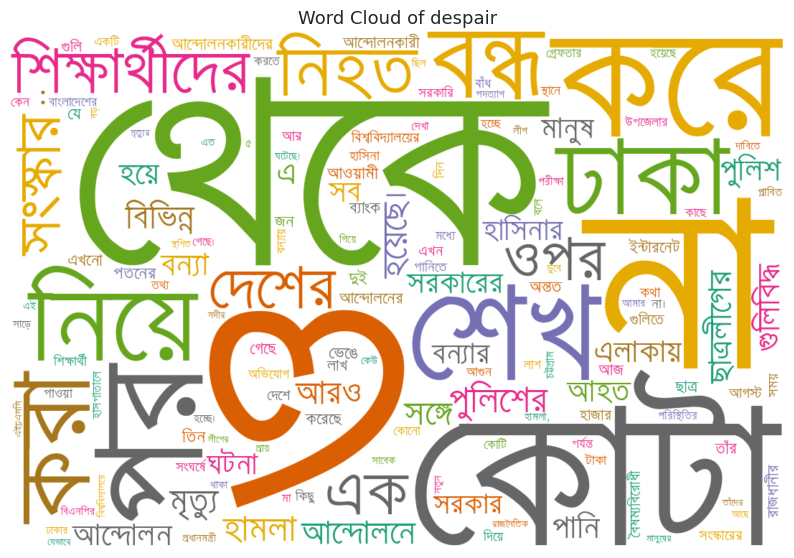} 
        \caption{Despair}
    \end{subfigure}
    \hfill
    \begin{subfigure}{0.32\columnwidth}
        \centering
        \includegraphics[width=\linewidth]{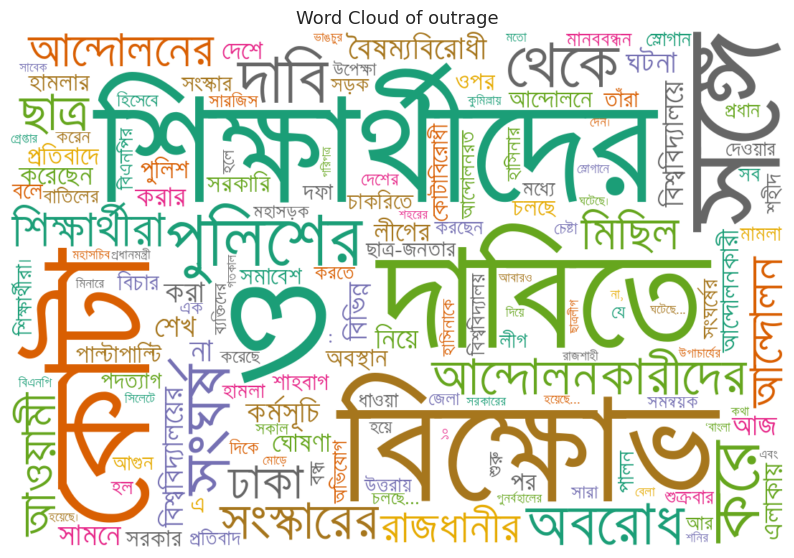} 
        \caption{Outrage}
    \end{subfigure}
    \hfill
    \begin{subfigure}{0.32\columnwidth}
        \centering
        \includegraphics[width=\linewidth]{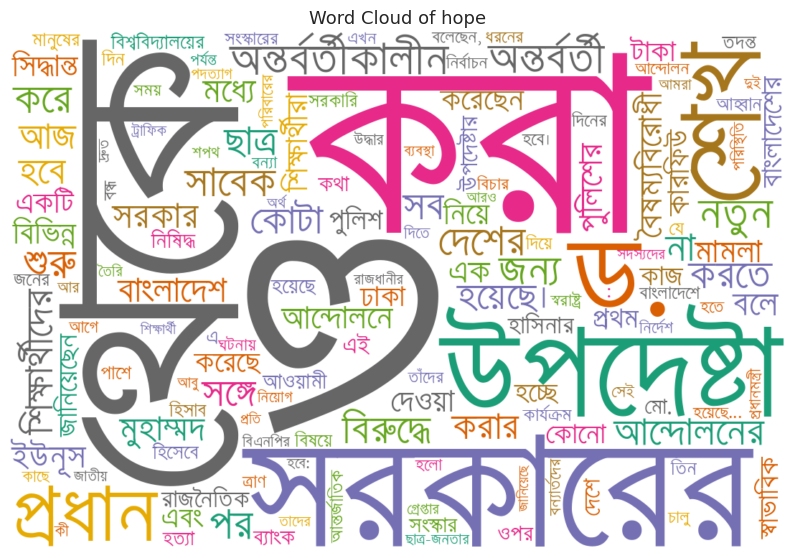} 
        \caption{Hope}
    \end{subfigure}
    \caption{Word clouds for the three sentiment classes.}
    \label{fig:word_clouds}
\end{figure}

We classify crisis-related emotions into three categories: Outrage ($\mathcal{O}$), Hope ($\mathcal{H}$), and Despair ($\mathcal{D}$), based on protest psychology \cite{jasper1998emotions}. Each headline $h_i$ is annotated by $A=\{a_1,a_2,a_3\}$, with majority voting yielding final labels:
\begin{equation}
    \hat{y}_i = \arg\max_{c \in \{\mathcal{O},\mathcal{H},\mathcal{D}\}} \sum_{j=1}^{3} \mathbb{I}[y_i^{(j)}=c] \cdot w_j.
\end{equation}
Cohen's kappa $\kappa = 0.78$ indicates substantial agreement on a 500-sample pilot. A five-class variant showed lower agreement, so a three-class taxonomy is retained for clarity and reliability. Figure~\ref{fig:workflow} shows the full annotation pipeline.Also Fig.~\ref{fig:word_clouds} shows word clouds for despair, hope, and outrage, highlighting the most frequent words and key themes within each sentiment

\begin{figure}[h]
    \centering
    \begin{subfigure}{0.48\columnwidth}
        \centering
        \includegraphics[width=\linewidth]{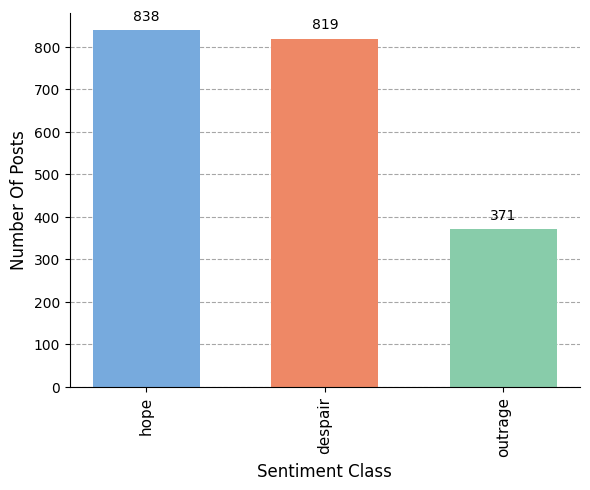}
        \caption{Class Distribution Before Augmentation}
        \label{fig:Class_Dist1}
    \end{subfigure}\hfill
    \begin{subfigure}{0.48\columnwidth}
        \centering
        \includegraphics[width=\linewidth]{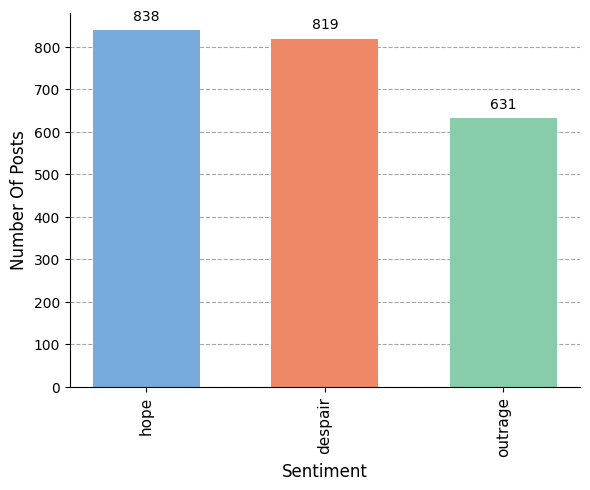}
        \caption{Class Distribution After Augmentation}
        \label{fig:Class_Dist2}
    \end{subfigure}
    \caption{Class distributions before and after augmentation.}
    \label{fig:class_distributions}
\end{figure}

\subsection{Data Augmentation Strategy}
To address the under-representation of the "Outrage" class, we applied paraphrasing-based augmentation with BanglaT5 \cite{bhattacharjee2022banglanlg} on the training set:
\begin{equation}
    \tilde{h}_i = \text{BanglaT5}(h_i;\, \theta_{\text{para}}), \quad h_i \in \mathcal{O}_{\text{train}}.
\end{equation}
This increased the number of Outrage samples to 631 (see Figure~\ref{fig:Class_Dist2} and Figure~\ref{fig:Class_Dist1} before and after augmentation of the dataset).

\begin{figure*}[t]
    \centering
    \includegraphics[width=0.70\textwidth]{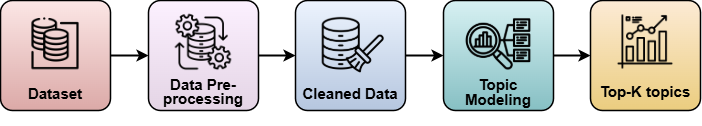}
    \caption{Topic Modeling using LDA to identify K topics from the dataset.}
    \label{fig:lda_workflow}
\end{figure*}

\subsection{Linguistic Preprocessing Pipeline}
\subsubsection{Punctuation Removal}
We removed punctuation and special characters to clean the text, ensuring a consistent format for further processing. For example:
\begin{quote}
 \bn{কোটাবিরোধী আন্দোলন : ঢাকা বিশ্ববিদ্যালয়ে মিছিল শুরু, মধুর ক্যানটিনে জড়ো হয়েছে ছাত্রলীগ}
\end{quote}
becomes:
\begin{quote}
 \bn{কোটাবিরোধী আন্দোলন ঢাকা বিশ্ববিদ্যালয়ে মিছিল শুরু মধুর ক্যানটিনে জড়ো হয়েছে ছাত্রলীগ}
\end{quote}

\subsubsection{Tokenization}
Each headline $h$ is tokenized into a sequence of words, which facilitates downstream modeling:
\begin{quote}
 \bn{চট্টগ্রামে ট্রেন আটকে শিক্ষার্থীদের আন্দোলন} $\rightarrow$ [ \bn{চট্টগ্রামে}, \bn{ট্রেন}, \bn{আটকে}, \bn{শিক্ষার্থীদের}, \bn{আন্দোলন} ]
\end{quote}

\subsubsection{Bangla Stopword Removal}
We filtered out stopwords to emphasize sentiment-bearing terms:
\begin{quote}
 \bn{কোটাবিরোধী ও সর্বজনীন পেনশন প্রত্যাহারের আন্দোলনে নৈতিক সমর্থন দিল বিএনপি} $\rightarrow$ 
 \bn{কোটাবিরোধী সর্বজনীন পেনশন প্রত্যাহারের আন্দোলনে নৈতিক সমর্থন বিএনপি}
\end{quote}
using a publicly available stopword list \cite{afnan2023banglastopwords}.

\subsubsection{Bangla Stemming}
Morphological variants were reduced to root forms using a Bangla stemmer.

Examples:
\bn{চালিয়ে চালানো চালাবে} $\rightarrow$ \bn{চালা};\quad
\bn{আন্দোলনকারী, আন্দোলনকারীদের} $\rightarrow$ \bn{আন্দোলন}

\subsection{Stratified Partitioning Strategy}
We used stratified splits to ensure fair model evaluation. For traditional models (SVM, Logistic Regression), we reserved 30\% as a test set with a 70/30 split. Deep learning models (BanglaBERT, XLM-RoBERTa) were split into 70\% training, 15\% validation, and 15\% test. To address class imbalance in "Outrage," we applied paraphrasing augmentation, improving class distribution and ensuring fair evaluation.


\section{Topic Modeling with LDA}
We apply Latent Dirichlet Allocation (LDA) \cite{blei2003latent} to identify latent topics in news headlines from the July–August 2024 mass uprising in Bangladesh. Beyond the full corpus, we analyze sentiment-specific subsets—outrage, despair, and hope—to capture both broad and emotion-specific themes. As shown in Fig.\ref{fig:lda_workflow}, the workflow includes preprocessing, cleaning, LDA modeling, and extraction of K topics. This pipeline enables a focused exploration of the thematic patterns and emotional undertones present in the news coverage.

\subsection{Preliminaries}
LDA is a generative probabilistic model that models documents as mixtures of latent topics, where each topic is characterized by a multinomial distribution over words \cite{blei2003latent}. Given a corpus with $M$ documents, vocabulary size $V$, and $K$ topics, LDA assumes each document $i$ with $N_i$ words is generated through a hierarchical Bayesian process.

\textbf{Mathematical Foundation:} Let $\theta_i$ be the topic proportions for document $i$, $\phi_k$ be the word distribution for topic $k$, and $z_{i,j}$ be the topic assignment for word $w_{i,j}$. The generative process follows:
\begin{enumerate}[leftmargin=*,nosep]
    \item For each topic $k$: Draw $\phi_k \sim \text{Dirichlet}(\beta)$, where $\beta$ controls topic sparsity
    \item For each document $i$: Draw $\theta_i \sim \text{Dirichlet}(\alpha)$, where $\alpha$ controls document-topic mixing
    \item For each word position $(i,j)$:
    \begin{itemize}[leftmargin=*,nosep]
        \item Sample topic: $z_{i,j} \sim \text{Multinomial}(\theta_i)$
        \item Sample word: $w_{i,j} \sim \text{Multinomial}(\phi_{z_{i,j}})$
    \end{itemize}
\end{enumerate}

The joint probability of observed data and latent variables is:
\begin{multline}
P(W, Z, \Theta, \Phi \mid \alpha, \beta) =
\prod_{k=1}^K p(\phi_k \mid \beta) \,
\prod_{i=1}^M p(\theta_i \mid \alpha) \\
\times \prod_{j=1}^{N_i} p(z_{i,j} \mid \theta_i) \, p(w_{i,j} \mid \phi_{z_{i,j}})
\end{multline}

\textbf{Inference Challenge:} The posterior $P(Z, \Theta, \Phi \mid W, \alpha, \beta)$ is intractable due to coupling between latent variables. We employ Variational Bayes with mean-field approximation:
\begin{equation}
q(\Theta, \Phi, Z) = \prod_{i=1}^{M} q(\theta_i; \gamma_i) \prod_{k=1}^{K} q(\phi_k; \lambda_k) \prod_{i,j} q(z_{i,j}; \varphi_{i,j})
\end{equation}
where $q(\theta_i)$ and $q(\phi_k)$ are Dirichlet distributions, and $q(z_{i,j})$ are categorical. Parameters are optimized by maximizing the Evidence Lower BOund (ELBO).

\textbf{Document Representation:} Each document $i$ is represented as a bag-of-words vector $\mathbf{c}_i \in \mathbb{N}^V$ where $c_{i,v} = |\{j : w_{i,j} = v\}|$ counts occurrences of vocabulary term $v$.

\textbf{Model Selection:} We select the optimal number of topics via coherence maximization:
\begin{equation}
K^{\star} = \arg\max_{K} C_v(K)
\end{equation}
yielding $K^{\star}=10$. For sentiment-specific analysis, we train separate LDA models on $\mathcal{D}_\text{Outrage}$, $\mathcal{D}_\text{Hope}$, and $\mathcal{D}_\text{Despair}$ using identical hyperparameters $(\alpha,\beta)$, enabling direct comparison of emotion-conditioned topic distributions.

\begin{figure*}[htbp]
    \centering
    \includegraphics[width=0.65\textwidth]{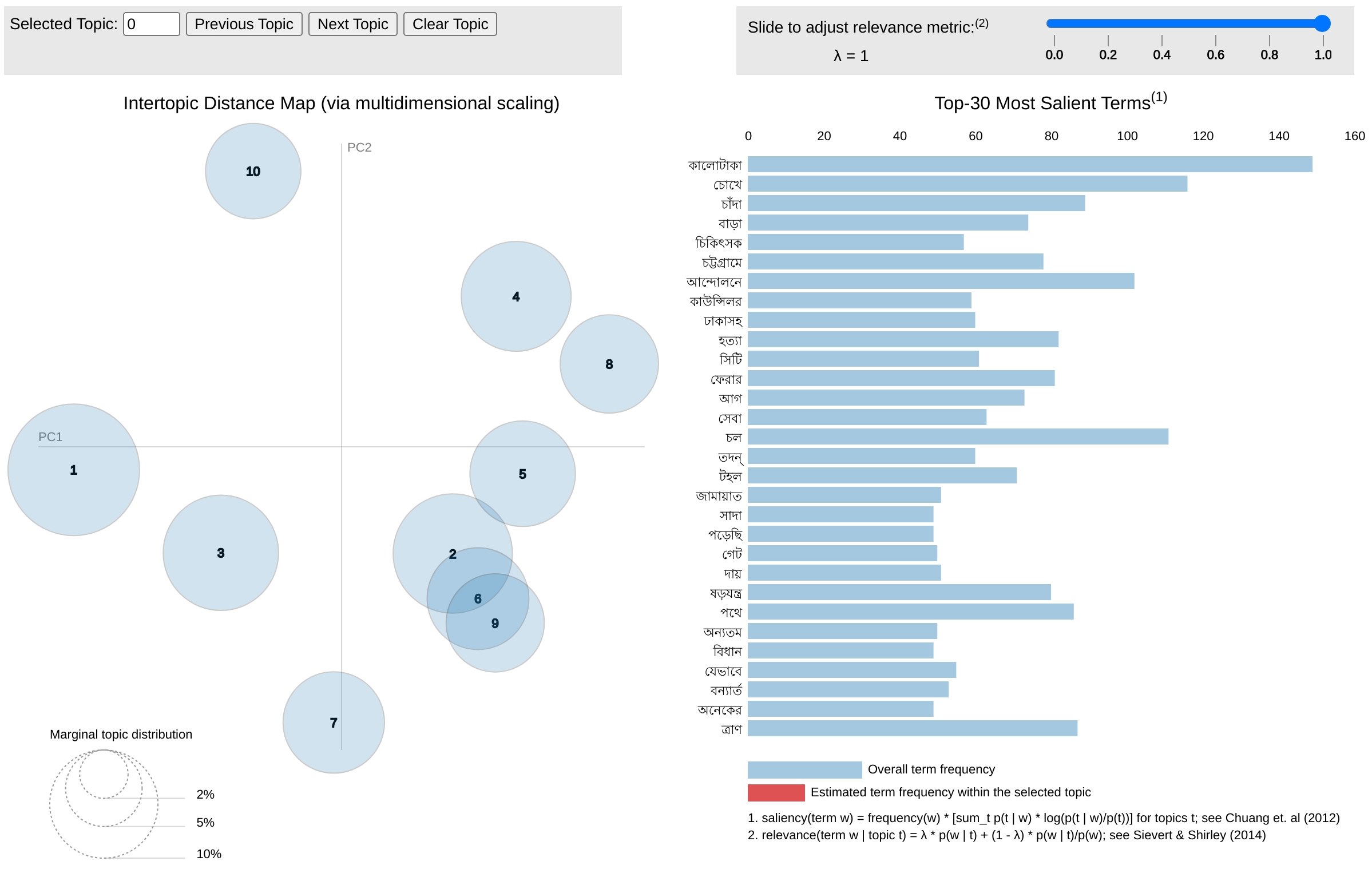}
    \caption{pyLDAvis visualizations for the full dataset LDA : Full Dataset}
    \label{fig:lda_full}
\end{figure*}


\subsection{Overall Topic Analysis}
Applying LDA to the full dataset revealed several dominant themes (see Fig~\ref{fig:lda_full}). Notable topics include:
\begin{itemize}[leftmargin=*,nosep]
    \item \textbf{Topic 0:} \textit{Financial Misconduct and Political Corruption} (keywords: \bn{কালোটাকা}, \bn{বাস}, \bn{নির্বাচন}, \bn{রাজনৈতিক}).
    \item \textbf{Topic 2:} \textit{Political Protests and Government Action} (keywords: \bn{আন্দোলন}, \bn{আওয়ামী}, \bn{সরক}, \bn{কোটা}).
    \item \textbf{Topic 6:} \textit{Crime and Police Activity} (keywords: \bn{হত্যা}, \bn{চট্টগ্রামে}, \bn{পুলিশ}, \bn{থানা}).
\end{itemize}
A coherence score of 0.5088 suggests that the topics are clear and well-defined, effectively capturing the broader issues of political corruption, governance, and public order during the uprising.

\subsection{Sentiment-Specific Topic Analysis}
In addition to the full dataset, we conducted separate LDA analyses on sentiment-specific subsets to explore the nuances of public emotion.

\textbf{Outrage:}  
The LDA model for the 'outrage' subset reveals topics heavily influenced by public anger and protest dynamics. Most topics (e.g., Topics 0, 2, 3, 4, 5, 6, 8, 9) focus on student-led protests, police confrontations, and political accusations (keywords: \bn{দাবি}, \bn{শিক্ষার্থী}, \bn{পুলিশ}, \bn{আন্দোলনকারী}, \bn{কোটা}). Despite a lower coherence score of 0.3954, these topics clearly reflect intense sentiments directed at government policies and public order issues, particularly concerning quota reforms.

\textbf{Despair:}  
The despair analysis captures the compounded effects of natural disasters and political unrest. For instance, one topic (Topic 0) emphasizes the impact of natural calamities (keywords: \bn{বন্যা}, \bn{পরিস্থিতির অবনতি}), while another (Topic 1) highlights violence during disasters (keywords: \bn{নদীর}, \bn{নিহত}, \bn{বাঁধ}, \bn{হামলা}). Additional topics (Topics 3 and 4) delve into violent clashes and systemic instability, leading to an overall coherence score of 0.4488. This suggests that a mixture of environmental and socio-political stressors contributed to a sense of hopelessness.

\textbf{Hope:}  
The 'hope' subset reveals topics that express cautious optimism about political change. Keywords such as \bn{উপদেষ্}, \bn{সরকার}, \bn{শেখ}, and \bn{অন্তর্বর্তীকালীন} indicate a public expectation for a caretaker government and leadership reforms. Although the coherence score is the lowest (0.3891), the topics suggest that even amid turmoil, there remains an undercurrent of hope for a more just and neutral administrative transition.

\begin{figure*}[t]
\centering
\includegraphics[width=\textwidth]{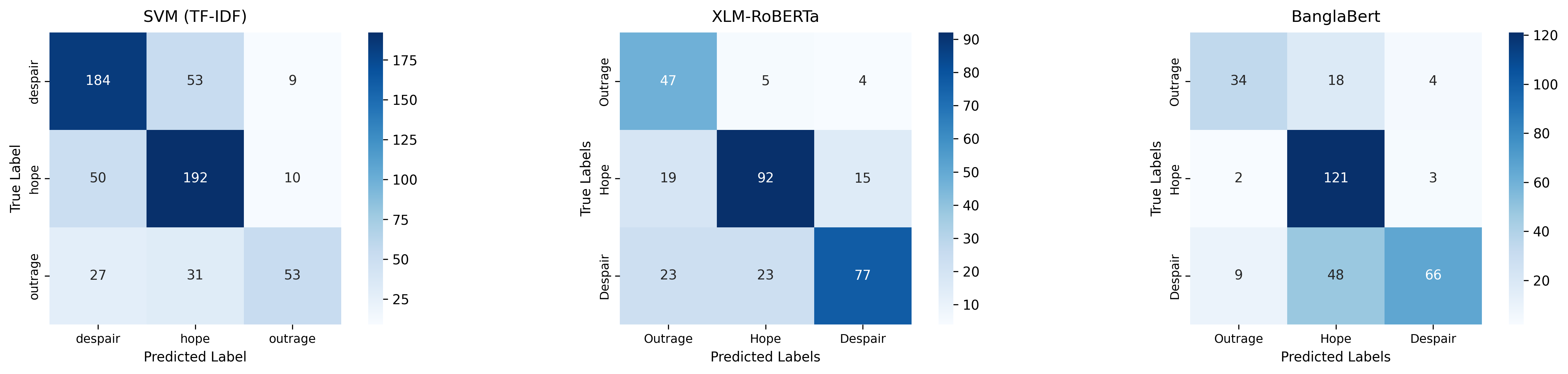}
\caption{Confusion Matrix for Language Models and ML Algorithms.}
\label{fig:cm}
\end{figure*}
\subsection{Analysis}
The LDA analyses provide a multi-layered perspective on the 2024 Bangladesh Uprising. The full dataset analysis outlines broad issues of political corruption, public protests, and law enforcement challenges. In contrast, sentiment-specific models highlight distinct emotional narratives:
\begin{itemize}[leftmargin=*,nosep]
    \item \textbf{Outrage:} Reflects active resistance and public discontent, particularly among student communities.
    \item \textbf{Despair:} Captures the interplay between environmental disasters and political violence, deepening the sense of crisis.
    \item \textbf{Hope:} Indicates public aspirations for political reform and a transitional government.
\end{itemize}
While the overall model achieves relatively high coherence, the sentiment-specific analyses—despite lower coherence scores offer valuable insights into the nuanced emotional drivers behind the public discourse. Future work may focus on parameter tuning to improve clarity in these more granular models.

An example Bangla headline illustrating these dynamics is:
\begin{quote}
 \bn{কোটাবিরোধী আন্দোলন ঢাকা বিশ্ববিদ্যালয়ে মিছিল শুরু মধুর ক্যানটিনে জড়ো হয়েছে ছাত্রলীগ}
\end{quote}

\section{Experimental Setup and Results}
The experiments were conducted using Google Colab. It provides access to NVIDIA GPUs, such as the Tesla T4, which significantly accelerated the training and evaluation of machine learning models. This made it possible to experiment with large models like BanglaBERT, mBERT, and XLM-RoBERTa without the need for local infrastructure.

\begin{table}[h!]
\centering
\footnotesize 
\setlength{\tabcolsep}{10pt} 
\begin{tabular}{lcccc}
\hline
\textbf{Model} & \textbf{Acc.} & \textbf{Prec.} & \textbf{Rec.} & \textbf{F1} \\
\hline
\multicolumn{5}{c}{\textbf{Machine Learning}} \\
\hline
SVM & \textbf{0.70} & \textbf{0.71} & \textbf{0.66} & \textbf{0.68} \\
Logistic Reg. & 0.70 & 0.69 & 0.67 & 0.68 \\
Random Forest & 0.64 & 0.64 & 0.61 & 0.62 \\
KNN & 0.63 & 0.60 & 0.60 & 0.60 \\
\hline
\multicolumn{5}{c}{\textbf{Language Models}} \\
\hline
BanglaBERT & \textbf{0.72} & \textbf{0.77} & \textbf{0.70} & \textbf{0.71} \\
XLM-RoBERTa & 0.71 & 0.70 & 0.73 & 0.70 \\
mBERT & 0.67 & 0.65 & 0.65 & 0.64 \\
Bangla Electra & 0.39 & 0.37 & 0.39 & 0.34 \\
\hline
\end{tabular}
\caption{Performance of ML algorithms and language models}
\label{tab:merged_performance}
\end{table}

\begin{table}[ht]
\centering
\begin{tabular}{c c c c}
\hline
\textbf{LLMs} & \textbf{Metric} & \textbf{Zero-shot} \\ \hline
\multirow{4}{*}{\shortstack{GPT 4o Mini}} & Accuracy & 0.68   \\  
                              & Precision & 0.73 \\ 
                              & Recall & 0.68   \\ 
                              & F1-Score & 0.68   \\ \hline
\multirow{4}{*}{\shortstack{Gemini 2.0 \\ Flash \\ Experimental}}    & Accuracy & 0.68  \\  
                              & Precision & 0.73 \\ 
                              & Recall & 0.70   \\ 
                              & F1-Score & 0.67 \\ \hline
\multirow{4}{*}{\shortstack{DeepSeek-\\R1}}   & Accuracy & 0.74   \\  
                              & Precision & 0.77  \\ 
                              & Recall & 0.75  \\ 
                              & F1-Score & 0.75  \\ \hline
\end{tabular}
\caption{Performance Metrics of LLMs}
\label{tab:llms}
\end{table}

\subsection{Results Analysis}

Figure~\ref{fig:cm} presents the class-wise confusion matrices for both transformer-based and classical models, while the overall performance is summarized in Tables~\ref{tab:merged_performance} and~\ref{tab:llms}.

Among transformer-based approaches, \textbf{BanglaBERT} achieved the highest accuracy at \textbf{72.0\%}, with strong precision (\textbf{0.77}) and balanced recall (\textbf{0.70}), resulting in an F1-score of \textbf{0.71}. \textbf{XLM-RoBERTa} followed closely with \textbf{71.0\%} accuracy and slightly higher recall (\textbf{0.73}), while \textbf{mBERT} reached \textbf{67.0\%} accuracy with more modest precision and recall values (both \textbf{0.65}).

Classical machine learning models using TF–IDF bigrams (10K features) remained competitive. Both \textbf{Logistic Regression} and \textbf{SVM} achieved \textbf{70.0\%} accuracy, with SVM providing the best balance across metrics (precision \textbf{0.71}, recall \textbf{0.66}, F1-score \textbf{0.68}).

In zero-shot large language model (LLM) evaluation, \textbf{DeepSeek-R1} attained the highest accuracy at \textbf{74.0\%}, along with consistently strong precision (\textbf{0.77}), recall (\textbf{0.75}), and F1-score (\textbf{0.75}). \textbf{GPT-4o Mini} and \textbf{Gemini 2.0 Flash Experimental} both achieved \textbf{68.0\%} accuracy, with precision around \textbf{0.73} and balanced recall–F1 values.

\section{Discussion}
Our experiments show that both transformer-based and classical models perform competitively on the Bangla political sentiment dataset, with accuracies ranging from 70\% to 72\%. BanglaBERT achieved the highest score (\textbf{72\%}), highlighting the benefits of language-specific pretraining, while Logistic Regression and SVM (\textbf{70\%}) demonstrated that well-engineered classical approaches can be similarly effective in resource-constrained settings. LDA analysis revealed that political corruption, mass protests, and law enforcement actions dominated public discourse, with temporal trends showing distinct sentiment shifts: a spike in \textbf{Despair} before the internet blackout (July 15--18), reduced \textbf{Outrage} during the blackout (July 19--23), a post-blackout rise in both \textbf{Hope} and \textbf{Despair} (July 24--31), an \textbf{Outrage} surge in August, increased \textbf{Hope} following the government's fall (Aug 5), and renewed \textbf{Despair} from late-August flooding. These findings underscore the interplay between political events and public emotions, offering valuable insights for policymaking and crisis communication strategies.

\section{Conclusion}
This study advances Bangla sentiment analysis by introducing a 2,028-headline dataset on the 2024 Bangladesh mass uprising, addressing the scarcity of resources for low-resource languages in socio-political contexts. Collected from a major Facebook news page with rigorous preprocessing, the dataset supports reliable downstream analysis. LDA revealed themes such as political corruption, protests, and environmental crises, offering policymakers insight into sentiment dynamics. Temporal trends showed despair rising during internet blackouts and disasters, and hope increasing after political transitions. However, reliance on Facebook headlines, short-text format, and a three-class emotion scheme limit broader generalization.


\subsection{Dataset Availability}
The dataset used in this research is publicly available on Kaggle: \href{https://www.kaggle.com/datasets/sami346/mass-uprising-2024-in-bangladesh}{Kaggle
 Dataset Link}.

\small
\bibliographystyle{IEEEtran}
\bibliography{bib} 

\end{document}